\documentclass[11pt,a4paper]{article}
\usepackage[hyperref]{acl2021}
\usepackage{times}
\usepackage{latexsym}
\usepackage{titlesec}
\usepackage{amsmath}
\usepackage{color,soul}
\usepackage{graphicx}
\graphicspath{{./images/}}
\usepackage{booktabs}
\usepackage{array}
\usepackage{float}
\usepackage{scrextend}

\usepackage{microtype}

\aclfinalcopy 

\makeatletter
\renewcommand\paragraph{%
    \@startsection{paragraph}{4}{0mm}%
        {-\baselineskip}%
        {.5\baselineskip}%
        {\normalfont\normalsize\bfseries}}
\makeatother

\title{BigGreen at SemEval-2021 Task 1: \\
Lexical Complexity Prediction with Assembly Models}

\author{
  \textbf{Aadil Islam}\normalfont{,} \textbf{Weicheng Ma}\normalfont{,} \textbf{Soroush Vosoughi}\\
  Department of Computer Science\\
  Dartmouth College\\
  \texttt{\{aadil.islam.21,weicheng.ma.gr,soroush.vosoughi\}@dartmouth.edu}
}

\date{}

\begin{document}
\maketitle

\begin{abstract}
  This paper describes a system submitted by team BigGreen to LCP 2021 for predicting the lexical complexity of English words in a given context. We assemble a feature engineering-based model with a deep neural network model founded on BERT. While BERT itself performs competitively, our feature engineering-based model helps in extreme cases, eg. separating instances of easy and neutral difficulty. Our handcrafted features comprise a breadth of lexical, semantic, syntactic, and novel phonological measures. Visualizations of BERT attention maps offer insight into potential features that Transformers models may learn when fine-tuned for lexical complexity prediction. Our ensembled predictions score reasonably well for the single word subtask, and we demonstrate how they can be harnessed to perform well on the multi word expression subtask too.
\end{abstract}

\section{Introduction}

Lexical simplification (LS) is the task of replacing difficult words in text with simpler alternatives. It is relevant in reading comprehension, where early studies have shown infrequent words lead to more time spent by a reader fixated on it, and that ambiguity in a word's meaning adds to comprehension time \citep{rayner1986lexical}. Complex word identification (CWI) is believed to be a fundamental step in the automation of lexical simplification \citep{shardlow2014out}. Early techniques for conducting CWI suffer from a lack of robustness, from simplifying all words to \textit{then} study its efficacy \citep{devlin1998use}, to applying thresholds on features like word frequency \citep{zeng2005text}. 

This year's Lexical Complexity Prediction (LCP) shared task \citep{shardlow2021semeval} forgoes the treatment of word difficulty as a binary classification task \citep{paetzold2016semeval, yimam2018report} and instead measures degree of complexity on a continuous scale. This choice is intriguing as it mitigates a dilemma with previous approaches of having to treat words extremely close to a decision boundary (suppose a threshold deems a word's difficulty) identically to those that are far away, ie. extremely easy or extremely difficult.

Teams are asked to submit predictions on unlabeled test sets for two subtasks: predicting on English single word and multi word expressions (MWEs). For each subtask, \texttt{BigGreen} presents a machine learning-based approach that fuses the predictions of a feature engineering-based regressor with those of a feature learning-based deep neural network model founded on BERT \citep{devlin2018bert}. Our code is made available on GitHub.\footnote{\url{https://github.com/Aadil101/BigGreen-at-LCP-2021}}

\section{Related Work}

Previous studies have looked at estimating the readability of a given text at the sentence-level. \citet{mc1969smog} regresses the number of polysyllabic words in a given lesson against the mean score for students quizzed on said lesson, yielding the SMOG Readability Formula. \citet{dale1948formula} offer a list of 768 (later updated to 3,000) words familiar to grade-school students in reading, which they find correlates with passage difficulty. An issue with traditional readability metrics seems to be the loss of generality at the word-level.

\citet{shardlow2013comparison} tries a brute force approach where a simplification algorithm is applied to each word of a given text, deeming a word complex only if it is simplified. However, this suffers from the assumption that a non-complex word does not require further simplification. They also try assigning a familiarity score to a word, and determining whether the word is complex or not by applying a threshold. We avoid thresholding our features in this study as we find it unnecessary, since raw familiarity scores can be used as features in regression-based tasks. 

Results from CWI at SemEval-2016 \citep{zampieri2017complex} suggest vote ensembling predictions of the best performing models as an effective strategy, while several top-performing models \citep{paetzold2016sv000gg, ronzano2016taln, mukherjee2016ju_nlp} appear to use linguistic information beyond just word frequency. This inspires our use of ensemble techniques, and a foray into phonological features as a new point of research. Results from CWI at SemEval-2018 show feature engineering-based models outperforming deep learning-based counterparts, despite the latter having generally better performances since SemEval-2016.

\section{Data}

\subsection{CompLex Dataset}

\begin{table}
  \centering
  \begin{tabular}{l|l|r|r|r}
    \hline
    \centering
    \textbf{Corpus} & \textbf{Subtask} & \textbf{Train} &  \textbf{Trial} &  \textbf{Test} \\
    \hline
    Bible & Single Word &   2574 &    143 &   283 \\
            & Multi Word &    505 &     29 &    66 \\
    \hline
    Biomed & Single Word &   2576 &    135 &   289 \\
            & Multi Word &    514 &     33 &    53 \\
    \hline
    Europarl & Single Word &   2512 &    143 &   345 \\
            & Multi Word &    498 &     37 &    65 \\
    \hline
  \end{tabular}
  \caption{\label{tab:datasets} LCP train, trial, and test sets.}
\end{table}

\citet{shardlow2020complex} present CompLex, a novel dataset in which each target expression (a single word or two-token MWE) is assigned a continuous label denoting its lexical complexity. Each label lies in range 0-1, and represents the (normalized) average score given by employed crowd workers who record an expression's difficulty on a 5-point Likert scale. We define a sample's \textit{class} as the bin to which its complexity label belongs, where bins are formed using the following mapping of complexity ranges: $[0,0.2) \rightarrow 1$, $[0.2, 0.4) \rightarrow 2$, $[0.4, 0.6) \rightarrow 3$, $[0.6, 0.8) \rightarrow 4$, $[0.8, 1] \rightarrow 5$. Target expressions in CompLex have 0.395 average complexity and 0.115 standard deviation, reflecting an imbalance in favor of class 2 and 3 samples. 

Each target expression is accompanied by the sentence it was extracted from, drawn from one of three corpora (Bible, Biomed, and Europarl). A summary of the train, trial,\footnote{In our study we avoid the trial set as we find it to be less representative of the training data, opting instead for training set cross-validation (stratified by corpus and complexity label).} and test set samples is provided in Table \ref{tab:datasets}.

\subsection{External Datasets}

We use four additional corpora to extract term frequency-based features from:

\begin{itemize}
  \item \noindent \textbf{English Gigaword Fifth Edition} (Gigaword): this comprises articles from seven English newswires \citep{parker2011english}.
  \item \noindent \textbf{Google Books Ngrams, version 2} (GBND): this is used to count occurences of phrases across a corpus of books, accessed via the PhraseFinder API \citep{phrasefinder}.
  \item \noindent \textbf{British National Corpus, version 3} (BNC): this is a collection of written and spoken English text \citep{bnc2007british}.
  \item \noindent \textbf{SUBTLEXus}: this consists of American English subtitles, offering a multitude of word frequency lists \citep{brysbaert2009moving}.
\end{itemize}

\section{BigGreen System \& Approaches}

In this section, we overview features fed to our feature engineering-based model, as well as training techniques for the feature learning-based model. We describe our features in detail in Appendix \ref{appendix:descriptions}. Note that fitted models for the single word subtask are then harnessed for the MWE subtask.

\subsection{Feature Engineering-based Approach}

\subsubsection{Feature Extraction}

We aim to capture a breadth of information pertaining to the target word and its context. Most features follow heavily right-skewed distributions, prompting us to also consider the $\log$-transformed version of each feature. For the MWE subtask, features are extracted independently for head and tail words. 

\paragraph{Lexical Features}

These are features based on lexical information about the target word:

\begin{itemize}
  \item \textbf{Word length}: length of the target word.
  \item \textbf{Number of syllables}: number of syllables in the target word, via the Syllables library.\footnote{\url{https://github.com/prosegrinder/python-syllables}}
  \item \textbf{Is acronym}: whether the target word is a sequence of capital letters.
\end{itemize}
  
\paragraph{Semantic Features}

These features capture the target word's meaning:

\begin{itemize}
  \item \textbf{WordNet features}: the number of hyponyms and hypernyms associated with the target word in WordNet \citep{fellbaum2010wordnet}.
  \item \textbf{GloVe word embeddings}: we extract 300-dimension embeddings pre-trained on Wikipedia-2014 and Gigaword \citep{pennington2014glove} for each (lowercased) target word. 
  \item \textbf{ELMo word embeddings}: we extract for each target word a 1024-dimension contextualized embedding pre-trained on the One Billion Word Benchmark \citep{peters2018deep}.
  \item \textbf{GloVe context embeddings}: we obtain the average 300-dimension GloVe word embedding over each token in the given sentence.
  \item \textbf{InferSent context embeddings}: we obtain 4096-dimension InferSent embeddings \citep{conneau2017supervised} for each sentence.
\end{itemize}

\paragraph{Phonetic Features}

These features compute the likelihood that soundable portions of the target word would arise in English language. We estimate ground truth transition probabilities between any two units (phonemes or characters) using Gigaword:

\begin{itemize}
  \item \textbf{Phoneme transition probability}: we consider the min/max/mean/standard deviation over the set of transition probabilities for the target word's phoneme bigrams.
  \item \textbf{Character transition probability}: analogous to that above, over \textit{character} bigrams.
\end{itemize}

\paragraph{Word Frequency \& N-gram Features}

These features are expressly included due to their expected importance as features \citep{zampieri2017complex}. Gigaword is the main corpus from which we extract word frequency measures (for both lemmatized and unlemmatized versions of the target word), summed frequency of the target word's byte pair encodings (BPEs), as well as summed frequencies of bigrams and trigrams. We complement these features with their IDF-based analogues. Lastly, we use the GBND, BNC, and SUBTLEXus corpora to extract secondary word frequency, bigram, and trigram measures. 

\paragraph{Syntactic Features}

These are features that assess the syntactic structure of the target word's context. We construct the constituency parse tree for each sentence using a Stanford CoreNLP pipeline \citep{manning2014stanford}.

\begin{itemize}
  \item \textbf{Part of speech (POS)}: tag is assigned using NLTK's \texttt{pos\_tag} method \citep{bird2009natural}.
  \item \textbf{Depth of parse tree}: the parse tree's height.
  \item \textbf{Depth of target word}: distance (in edges) between target word and parse tree's root node.
  \item \textbf{Is proper}: whether the target word is a proper noun/adjective, detected using capitalization.
\end{itemize}

\subsubsection{Training}

Prior to training, we Z-score standardize all features. For the single word subtask, we fit Linear, Lasso \citep{tibshirani1996regression}, Elastic Net \citep{zou2005regularization}, Support Vector Machine \citep{platt1999probabilistic}, K-Nearest Neighbors \citep{wiki:K-nearest_neighbors_algorithm}, and XGBoost \citep{chen2016xgboost} regression models. 

After identifying the best performing model by Pearson correlation, we seek to mitigate the imbalanced nature of the target variable, ie. multitude of class 1,2,3 and lack of class 4,5 samples: we devise a sister version of our top-performing model, fit upon a \textit{reduced} training set. For the \textit{reduced} set, we tune percentages removed from classes 1-3 by performing cross-validation on the full  training set.

\subsection{Approach based on Feature Learning}

Our handcrafted feature set relies heavily on target word-specific features. Beyond N-gram and syntactic features, it is a cursory analysis of the context surrounding the target word. We seek an alternative, automated approach using feature learning.

\subsubsection{Architecture}

LSTM-based approaches have been used to model the contexts of target words in past works \citep{hartmann2018nilc, de2018deep}. An issue with a single LSTM is its ability to read tokens of an input sentence sequentially only in a single direction (eg. left-to-right). It inspires us to try a Transformer-based approach \citep{vaswani2017attention}, architectures that process sentences as a whole (instead of word-by-word) by applying \textit{attention} mechanisms upon them. Attention weights are useful as they can be interpreted as learned relationships between words. BERT \citep{devlin2018bert} is one such model used for a variety of natural language understanding (NLU) tasks.

Multi-Task Deep Neural Network (MT-DNN) proposed by \citet{liu2019multi} offers state-of-the-art results for multiple NLU tasks by incorporating benefits of both multi-task learning and language model pre-training. We are able to initialize MT-DNN's shared text encoding layers with a pre-trained BERT base model (cased), and fine-tune its later layers for 5 epochs, using a mean squared error loss function and default hyperparameters. Such hyperparameter settings are provided in Appendix \ref{appendix:hyperparameters}. Note that the model is fine-tuned on only the CompLex corpus.

\subsubsection{Input Layer}

Data is fed to the model's input layer in \textit{PremiseAndOneHypothesis} format, premise and hypothesis being sentence and target word/MWE, respectively. The data is preprocessed by a BERT tokenizer, backed by Hugging Face \citep{wolf2020transformers}.

\subsubsection{Output Layer}

Our model's output layer produces the predicted lexical complexity for a given target word/MWE. Additionally, we extract \textit{attention maps} across each of the model's attention heads, for each test set sample. These will be assessed in Section \ref{sec:bert_attention}.

\subsection{Ensembling}

\label{sec:ensembling}

Our best performing feature engineering-based regression model yields two sets of predictions (from fitting on \textit{full} and \textit{reduced} training sets, respectively). We default to using the \textit{full} predictions, then tune a threshold, where predictions higher than the threshold (likely of class 4,5 samples) are overwritten with the \textit{reduced} predictions. We compute a weighted average ensemble of these predictions with those of our MT-DNN model to obtain a final set of predictions for the single word subtask. 

For the MWE subtask, the fitted models from the previous subtask are harnessed to predict lexical complexities for the head and tail words. We then compute a weighted average ensemble of these predicted complexities \textit{and} the predictions of an MT-DNN model trained on MWEs.

\section{Results}

\begin{table}
  \centering
  \begin{tabular}{lcccc}
  \hline \textbf{Model} & \textbf{Pearson} & \textbf{Rank} \\ \hline
  XGBoost$_\textit{full}$ &	0.7589 & - \\
  XGBoost$_\textit{reduced}$ &	0.7456 & - \\
  XGBoost$_{\textit{full}+\textit{reduced}}$ & 0.7576 & - \\
  MT-DNN & 0.7484 & - \\
  Ensemble (submission) & 0.7749 & 8/54 \\
  \hline
  Best competition results & 0.7886 & \\ 
  \hline
  \end{tabular}
  \caption{\label{tab:single-word-results} Test set results for single word subtask.}
\end{table}

\begin{table}
  \centering
  \begin{tabular}{lcccc}
  \hline \textbf{Model} & \textbf{Pearson} & \textbf{Rank} \\ \hline
  XGBoost$_{\textit{full}+\textit{red.}}(\text{head})$ & 0.7164 & - \\
  XGBoost$_{\textit{full}+\textit{red.}}(\text{tail})$ & 0.7188 & - \\
  MT-DNN & 0.7890 & - \\
  Ensemble (submission) & 0.7898 & 25/37 \\
  Ensemble (improved) & 0.8290 & *14/37 \\
  \hline
  Best competition results & 0.8612 & \\ 
  \hline
  \end{tabular}
  \caption{\label{tab:multi-word-results} MWE subtask test set results. (*projection)}
\end{table}

We present the performances of \texttt{BigGreen}'s system on each subtask in Tables \ref{tab:single-word-results} and \ref{tab:multi-word-results}.

\section{Analysis}

\subsection{Performance}

\label{sec:performance}

For feature selection, we find success in selecting the top-300 features by mutual information and removing quasi-constant features. The pruned feature set is passed to wrapper/embedded methods and a variety of regressors for model comparison. We find an XGBoost regressor (with hyperparameters tuned via grid search) to excel consistently for the single word subtask. As shown in Table \ref{tab:single-word-results}, we rank in the top 15\% by Pearson correlation.

For the MWE subtask, performances are reported in Table \ref{tab:multi-word-results}. Note that our submitted predictions differ from post-competition predictions. We \textit{previously} used a training procedure resembling that for the single word subtask: (1) filter methods for feature selection, (2) XGBoost for regression, (3) ensembling with MT-DNN. We had passed the entire MWE as input to our XGBoost and MT-DNN models. We hypothesize that the fewer number of training samples available for this subtask contributed to the prior procedure's lackluster performance. This inspired us to incorporate the predictive capabilities of our fitted single word subtask models by applying them \textit{independently} on the MWE's constituent head and tail words. This gives us predicted complexities for the head and tail words each, which when ensembled with the predictions of our MT-DNN model (that, mind you, is trained on the \textit{entire} MWE) yields superior results to those submitted to competition.

\subsection{Feature Contribution}

\begin{figure}
  \centering
  \includegraphics[scale=0.3]{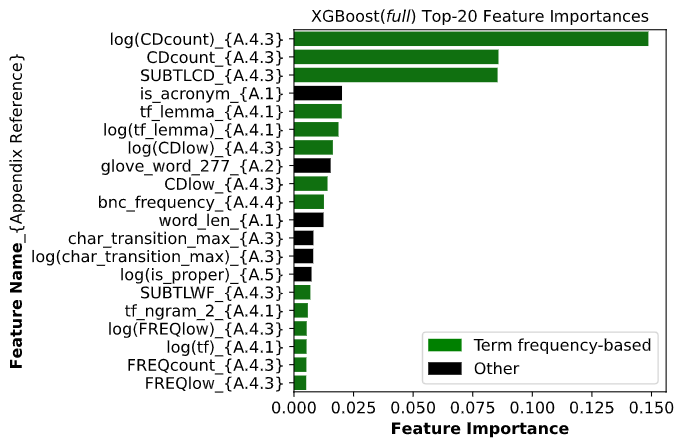}
  \captionsetup{justification=centering}
  \caption{\label{fig:xgboost_feature_importance} Feature importances for XGBoost$_\textit{full}$. Definitions of the features are shown in Appendix \ref{appendix:descriptions}.}
\end{figure}

In total we consider 110 features, in addition to our multidimensional embedding-based features and $\log$-transformed features. We inspect the estimated feature importance scores produced by the XGBoost$_\textit{full}$ model to find that term frequency-based features (eg. unigrams, bigrams, trigrams) are of overwhelming importance (see Figure \ref{fig:xgboost_feature_importance}). This raises concern for whether the MT-DNN model too relies on term frequencies to make \textit{its} predictions, and if not, the linguistic features it may have learned upon fine-tuning. Of the remaining features having non-zero feature importances, most appear to be dimensions of target word-based semantic features (ie. GloVe or ELMo embeddings).

\subsection{BERT Attention}
\label{sec:bert_attention}

\begin{figure}
  \centering
  \includegraphics[scale=0.37]{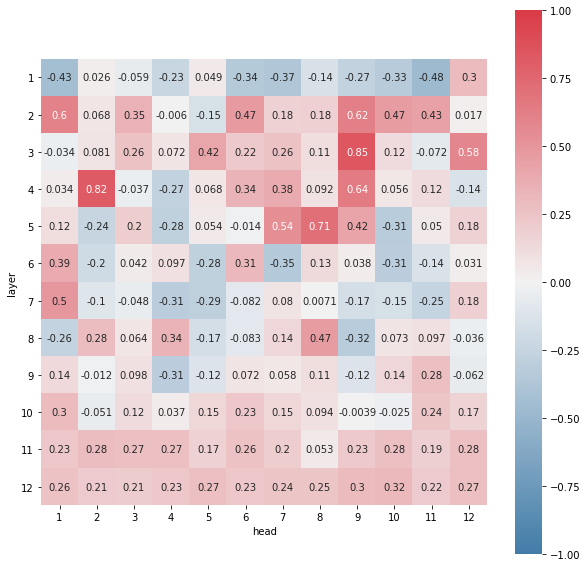}
  \captionsetup{justification=centering}
  \caption{\label{fig:head_correlations_tf} Attention head correlation between word frequency and total attention received by word, averaged across 100 random test set samples.}
\end{figure}

\begin{figure}
  \centering
  \includegraphics[scale=0.44]{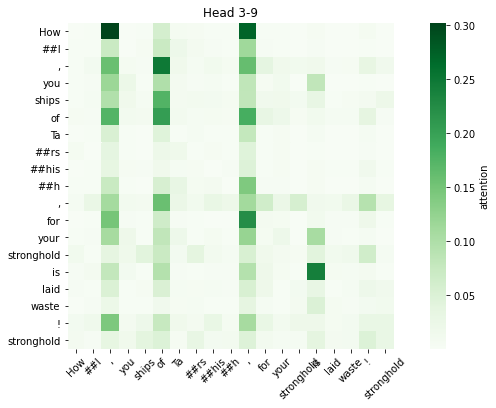}
  \captionsetup{justification=centering}
  \caption{\label{fig:head_3-9} Head 3-9 attention map for a random sample.}
\end{figure}

Attention maps have in previous works been assessed to demonstrate linguistic phenomena learned by a Transformer's specialized attention heads \citep{voita2019analyzing, clark2019does}. We extract attention maps from MT-DNN's underlying fine-tuned BERT architecture. For each sample in the single word test set, we obtain an attention map from each of the BERT base model's 144 attention heads (ie. 12 heads per 12 layers).

Based on the precedence given to term frequency features by the XGBoost$_\textit{full}$ model, we hypothesize that for certain attention heads, the degree to which BPEs attend to one another varies relative to their word's rarity in lexicon. This follows the findings of \citet{voita2019analyzing}, who identify heads in which lesser frequent tokens are attended to semi-uniformly by a majority of sentence tokens. 

To test our hypothesis, we estimate for each attention head the Pearson correlation between word frequency and average attention given to each word in the context.\footnote{We compute attention given to a \textit{word} as the sum of attention given to its constituent BPEs. We use the GBND corpus to extract word frequencies, though any large corpora would suffice.} As illustrated in Figure \ref{fig:head_correlations_tf}, we find multiple attention heads appearing to specialize at directing attention towards the most or least frequent words (depending on sign of the correlation). Vertical stripe patterns like that in Figure \ref{fig:head_3-9} emerge as a result of attention originating from a spectrum of tokens. The findings seem to affirm the fundamental relevancy of word frequency to lexical complexity prediction, corroborating our intuition.

\section{Conclusion}

In this paper, we report inspirations for a system submitted by \texttt{BigGreen} to LCP SharedTask 2021, performing reasonably well for the single word subtask by adapting ensemble methods upon feature engineering and feature learning-based models. We see potential in future deep learning approaches, acknowledging the need for complementary word frequency-based handcrafted features for the time being. We surpass our submitted results for the MWE subtask, by utilizing the predictive capabilities of our single word subtask models.

Avenues for improvement include better data aggregation, as a relative lack of class 4,5 samples seems to hurt Pearson correlation across extremely complex samples. An approach may involve synthetic data generation using SMOGN \citep{branco2017smogn}. \citet{shardlow2020complex} acknowledge a reader's familiarity with a genre may affect perceived word complexity. However, the CompLex dataset lacks information on each annotator's expertise or background, which may offer valuable new insights.

\bibliographystyle{acl_natbib}
\bibliography{anthology,acl2021}

\begin{thebibliography}{41}
\expandafter\ifx\csname natexlab\endcsname\relax\def\natexlab#1{#1}\fi

\bibitem[{Bird et~al.(2009)Bird, Klein, and Loper}]{bird2009natural}
Steven Bird, Ewan Klein, and Edward Loper. 2009.
\newblock \emph{Natural language processing with Python: analyzing text with
  the natural language toolkit}.
\newblock O'Reilly Media, Inc.

\bibitem[{Branco et~al.(2017)Branco, Torgo, and Ribeiro}]{branco2017smogn}
Paula Branco, Lu{\'\i}s Torgo, and Rita~P Ribeiro. 2017.
\newblock Smogn: a pre-processing approach for imbalanced regression.
\newblock In \emph{First international workshop on learning with imbalanced
  domains: Theory and applications}, pages 36--50. PMLR.

\bibitem[{Brysbaert and New(2009)}]{brysbaert2009moving}
Marc Brysbaert and Boris New. 2009.
\newblock Moving beyond ku{\v{c}}era and francis: A critical evaluation of
  current word frequency norms and the introduction of a new and improved word
  frequency measure for american english.
\newblock \emph{Behavior research methods}, 41(4):977--990.

\bibitem[{Chen and Guestrin(2016)}]{chen2016xgboost}
Tianqi Chen and Carlos Guestrin. 2016.
\newblock Xgboost: A scalable tree boosting system.
\newblock In \emph{Proceedings of the 22nd acm sigkdd international conference
  on knowledge discovery and data mining}, pages 785--794.

\bibitem[{Clark et~al.(2019)Clark, Khandelwal, Levy, and
  Manning}]{clark2019does}
Kevin Clark, Urvashi Khandelwal, Omer Levy, and Christopher~D Manning. 2019.
\newblock What does bert look at? an analysis of bert's attention.
\newblock \emph{arXiv preprint arXiv:1906.04341}.

\bibitem[{Conneau et~al.(2017)Conneau, Kiela, Schwenk, Barrault, and
  Bordes}]{conneau2017supervised}
Alexis Conneau, Douwe Kiela, Holger Schwenk, Loic Barrault, and Antoine Bordes.
  2017.
\newblock Supervised learning of universal sentence representations from
  natural language inference data.
\newblock \emph{arXiv preprint arXiv:1705.02364}.

\bibitem[{Consortium et~al.(2007)}]{bnc2007british}
BNC Consortium et~al. 2007.
\newblock British national corpus.
\newblock \emph{Oxford Text Archive Core Collection}.

\bibitem[{Dale and Chall(1948)}]{dale1948formula}
Edgar Dale and Jeanne~S Chall. 1948.
\newblock A formula for predicting readability: Instructions.
\newblock \emph{Educational research bulletin}, pages 37--54.

\bibitem[{De~Hertog and Tack(2018)}]{de2018deep}
Dirk De~Hertog and Ana{\"\i}s Tack. 2018.
\newblock Deep learning architecture for complexword identification.
\newblock In \emph{Thirteenth Workshop of Innovative Use of NLP for Building
  Educational Applications}, pages 328--334. Association for Computational
  Linguistics (ACL); New Orleans, Louisiana.

\bibitem[{Devlin et~al.(2018)Devlin, Chang, Lee, and
  Toutanova}]{devlin2018bert}
Jacob Devlin, Ming-Wei Chang, Kenton Lee, and Kristina Toutanova. 2018.
\newblock Bert: Pre-training of deep bidirectional transformers for language
  understanding.
\newblock \emph{arXiv preprint arXiv:1810.04805}.

\bibitem[{Devlin(1998)}]{devlin1998use}
Siobhan Devlin. 1998.
\newblock The use of a psycholinguistic database in the simplification of text
  for aphasic readers.
\newblock \emph{Linguistic databases}.

\bibitem[{Fellbaum(2010)}]{fellbaum2010wordnet}
Christiane Fellbaum. 2010.
\newblock Wordnet.
\newblock In \emph{Theory and applications of ontology: computer applications},
  pages 231--243. Springer.

\bibitem[{Hartmann and Dos~Santos(2018)}]{hartmann2018nilc}
Nathan Hartmann and Leandro~Borges Dos~Santos. 2018.
\newblock Nilc at cwi 2018: Exploring feature engineering and feature learning.
\newblock In \emph{Proceedings of the Thirteenth Workshop on Innovative Use of
  NLP for Building Educational Applications}, pages 335--340.

\bibitem[{Lesk(1986)}]{lesk1986automatic}
Michael Lesk. 1986.
\newblock Automatic sense disambiguation using machine readable dictionaries:
  how to tell a pine cone from an ice cream cone.
\newblock In \emph{Proceedings of the 5th annual international conference on
  Systems documentation}, pages 24--26.

\bibitem[{Liu et~al.(2019)Liu, He, Chen, and Gao}]{liu2019multi}
Xiaodong Liu, Pengcheng He, Weizhu Chen, and Jianfeng Gao. 2019.
\newblock Multi-task deep neural networks for natural language understanding.
\newblock \emph{arXiv preprint arXiv:1901.11504}.

\bibitem[{Manning et~al.(2014)Manning, Surdeanu, Bauer, Finkel, Bethard, and
  McClosky}]{manning2014stanford}
Christopher~D Manning, Mihai Surdeanu, John Bauer, Jenny~Rose Finkel, Steven
  Bethard, and David McClosky. 2014.
\newblock The stanford corenlp natural language processing toolkit.
\newblock In \emph{Proceedings of 52nd annual meeting of the association for
  computational linguistics: system demonstrations}, pages 55--60.

\bibitem[{Mc~Laughlin(1969)}]{mc1969smog}
G~Harry Mc~Laughlin. 1969.
\newblock Smog grading-a new readability formula.
\newblock \emph{Journal of reading}, 12(8):639--646.

\bibitem[{Mukherjee et~al.(2016)Mukherjee, Patra, Das, and
  Bandyopadhyay}]{mukherjee2016ju_nlp}
Niloy Mukherjee, Braja~Gopal Patra, Dipankar Das, and Sivaji Bandyopadhyay.
  2016.
\newblock Ju\_nlp at semeval-2016 task 11: Identifying complex words in a
  sentence.
\newblock In \emph{Proceedings of the 10th International Workshop on Semantic
  Evaluation (SemEval-2016)}, pages 986--990.

\bibitem[{Paetzold and Specia(2016{\natexlab{a}})}]{paetzold2016semeval}
Gustavo Paetzold and Lucia Specia. 2016{\natexlab{a}}.
\newblock Semeval 2016 task 11: Complex word identification.
\newblock In \emph{Proceedings of the 10th International Workshop on Semantic
  Evaluation (SemEval-2016)}, pages 560--569.

\bibitem[{Paetzold and Specia(2016{\natexlab{b}})}]{paetzold2016sv000gg}
Gustavo Paetzold and Lucia Specia. 2016{\natexlab{b}}.
\newblock Sv000gg at semeval-2016 task 11: Heavy gauge complex word
  identification with system voting.
\newblock In \emph{Proceedings of the 10th International Workshop on Semantic
  Evaluation (SemEval-2016)}, pages 969--974.

\bibitem[{Parker et~al.(2011)Parker, Graff, Kong, Chen, and
  Maeda}]{parker2011english}
R~Parker, D~Graff, J~Kong, K~Chen, and K~Maeda. 2011.
\newblock English gigaword fifth edition ldc2011t07 (tech. rep.).
\newblock Technical report, Technical Report. Linguistic Data Consortium,
  Philadelphia.

\bibitem[{Pedregosa et~al.(2011)Pedregosa, Varoquaux, Gramfort, Michel,
  Thirion, Grisel, Blondel, Prettenhofer, Weiss, Dubourg
  et~al.}]{pedregosa2011scikit}
Fabian Pedregosa, Ga{\"e}l Varoquaux, Alexandre Gramfort, Vincent Michel,
  Bertrand Thirion, Olivier Grisel, Mathieu Blondel, Peter Prettenhofer, Ron
  Weiss, Vincent Dubourg, et~al. 2011.
\newblock Scikit-learn: Machine learning in python.
\newblock \emph{the Journal of machine Learning research}, 12:2825--2830.

\bibitem[{Pennington et~al.(2014)Pennington, Socher, and
  Manning}]{pennington2014glove}
Jeffrey Pennington, Richard Socher, and Christopher~D Manning. 2014.
\newblock Glove: Global vectors for word representation.
\newblock In \emph{Proceedings of the 2014 conference on empirical methods in
  natural language processing (EMNLP)}, pages 1532--1543.

\bibitem[{Peters et~al.(2018)Peters, Neumann, Iyyer, Gardner, Clark, Lee, and
  Zettlemoyer}]{peters2018deep}
Matthew~E Peters, Mark Neumann, Mohit Iyyer, Matt Gardner, Christopher Clark,
  Kenton Lee, and Luke Zettlemoyer. 2018.
\newblock Deep contextualized word representations.
\newblock \emph{arXiv preprint arXiv:1802.05365}.

\bibitem[{Platt et~al.(1999)}]{platt1999probabilistic}
John Platt et~al. 1999.
\newblock Probabilistic outputs for support vector machines and comparisons to
  regularized likelihood methods.
\newblock \emph{Advances in large margin classifiers}, 10(3):61--74.

\bibitem[{Rayner and Duffy(1986)}]{rayner1986lexical}
Keith Rayner and Susan~A Duffy. 1986.
\newblock Lexical complexity and fixation times in reading: Effects of word
  frequency, verb complexity, and lexical ambiguity.
\newblock \emph{Memory \& cognition}, 14(3):191--201.

\bibitem[{Ronzano et~al.(2016)Ronzano, Anke, Saggion et~al.}]{ronzano2016taln}
Francesco Ronzano, Luis~Espinosa Anke, Horacio Saggion, et~al. 2016.
\newblock Taln at semeval-2016 task 11: Modelling complex words by contextual,
  lexical and semantic features.
\newblock In \emph{Proceedings of the 10th International Workshop on Semantic
  Evaluation (SemEval-2016)}, pages 1011--1016.

\bibitem[{Shardlow(2013)}]{shardlow2013comparison}
Matthew Shardlow. 2013.
\newblock A comparison of techniques to automatically identify complex words.
\newblock In \emph{51st Annual Meeting of the Association for Computational
  Linguistics Proceedings of the Student Research Workshop}, pages 103--109.

\bibitem[{Shardlow(2014)}]{shardlow2014out}
Matthew Shardlow. 2014.
\newblock Out in the open: Finding and categorising errors in the lexical
  simplification pipeline.
\newblock In \emph{LREC}, pages 1583--1590.

\bibitem[{Shardlow et~al.(2020)Shardlow, Cooper, and
  Zampieri}]{shardlow2020complex}
Matthew Shardlow, Michael Cooper, and Marcos Zampieri. 2020.
\newblock Complex: A new corpus for lexical complexity predicition from likert
  scale data.
\newblock In \emph{Proceedings of the 1st Workshop on Tools and Resources to
  Empower People with REAding DIfficulties (READI)}.

\bibitem[{Shardlow et~al.(2021)Shardlow, Evans, Paetzold, and
  Zampieri}]{shardlow2021semeval}
Matthew Shardlow, Richard Evans, Gustavo Paetzold, and Marcos Zampieri. 2021.
\newblock Semeval-2021 task 1: Lexical complexity prediction.
\newblock In \emph{Proceedings of the 14th International Workshop on Semantic
  Evaluation (SemEval-2021)}.

\bibitem[{Tibshirani(1996)}]{tibshirani1996regression}
Robert Tibshirani. 1996.
\newblock Regression shrinkage and selection via the lasso.
\newblock \emph{Journal of the Royal Statistical Society: Series B
  (Methodological)}, 58(1):267--288.

\bibitem[{Trenkmann()}]{phrasefinder}
Martin Trenkmann.
\newblock {PhraseFinder} -- search millions of books for language use.
\newblock \url{https://phrasefinder.io}.
\newblock Accessed: 2021-02-08.

\bibitem[{Vaswani et~al.(2017)Vaswani, Shazeer, Parmar, Uszkoreit, Jones,
  Gomez, Kaiser, and Polosukhin}]{vaswani2017attention}
Ashish Vaswani, Noam Shazeer, Niki Parmar, Jakob Uszkoreit, Llion Jones,
  Aidan~N Gomez, Lukasz Kaiser, and Illia Polosukhin. 2017.
\newblock Attention is all you need.
\newblock \emph{arXiv preprint arXiv:1706.03762}.

\bibitem[{Voita et~al.(2019)Voita, Talbot, Moiseev, Sennrich, and
  Titov}]{voita2019analyzing}
Elena Voita, David Talbot, Fedor Moiseev, Rico Sennrich, and Ivan Titov. 2019.
\newblock Analyzing multi-head self-attention: Specialized heads do the heavy
  lifting, the rest can be pruned.
\newblock \emph{arXiv preprint arXiv:1905.09418}.

\bibitem[{Wikipedia(2021)}]{wiki:K-nearest_neighbors_algorithm}
Wikipedia. 2021.
\newblock {K-nearest neighbors algorithm} --- {W}ikipedia{,} the free
  encyclopedia.
\newblock
  \url{http://en.wikipedia.org/w/index.php?title=K-nearest\%20neighbors\%20algorithm&oldid=1008084290}.
\newblock [Online; accessed 02-April-2021].

\bibitem[{Wolf et~al.(2020)Wolf, Chaumond, Debut, Sanh, Delangue, Moi, Cistac,
  Funtowicz, Davison, Shleifer et~al.}]{wolf2020transformers}
Thomas Wolf, Julien Chaumond, Lysandre Debut, Victor Sanh, Clement Delangue,
  Anthony Moi, Pierric Cistac, Morgan Funtowicz, Joe Davison, Sam Shleifer,
  et~al. 2020.
\newblock Transformers: State-of-the-art natural language processing.
\newblock In \emph{Proceedings of the 2020 Conference on Empirical Methods in
  Natural Language Processing: System Demonstrations}, pages 38--45.

\bibitem[{Yimam et~al.(2018)Yimam, Biemann, Malmasi, Paetzold, Specia,
  {\v{S}}tajner, Tack, and Zampieri}]{yimam2018report}
Seid~Muhie Yimam, Chris Biemann, Shervin Malmasi, Gustavo~H Paetzold, Lucia
  Specia, Sanja {\v{S}}tajner, Ana{\"\i}s Tack, and Marcos Zampieri. 2018.
\newblock A report on the complex word identification shared task 2018.
\newblock \emph{arXiv preprint arXiv:1804.09132}.

\bibitem[{Zampieri et~al.(2017)Zampieri, Malmasi, Paetzold, and
  Specia}]{zampieri2017complex}
Marcos Zampieri, Shervin Malmasi, Gustavo Paetzold, and Lucia Specia. 2017.
\newblock Complex word identification: Challenges in data annotation and system
  performance.
\newblock \emph{arXiv preprint arXiv:1710.04989}.

\bibitem[{Zeng et~al.(2005)Zeng, Kim, Crowell, and Tse}]{zeng2005text}
Qing Zeng, Eunjung Kim, Jon Crowell, and Tony Tse. 2005.
\newblock A text corpora-based estimation of the familiarity of health
  terminology.
\newblock In \emph{International Symposium on Biological and Medical Data
  Analysis}, pages 184--192. Springer.

\bibitem[{Zou and Hastie(2005)}]{zou2005regularization}
Hui Zou and Trevor Hastie. 2005.
\newblock Regularization and variable selection via the elastic net.
\newblock \emph{Journal of the royal statistical society: series B (statistical
  methodology)}, 67(2):301--320.

\end{thebibliography}
\clearpage
\appendix

\section{Feature Descriptions}
\label{appendix:descriptions}

Here, we describe in greater detail the various features that were experimented with for our feature engineering-based model. Note that while this discussion regards the single word subtask, for the MWE subtask we compute the same features but for each of the head and tail words, respectively.

\subsection{Lexical Features}

\texttt{word\_len}
\begin{itemize}
  \item Character length of the target word.
\end{itemize}
\texttt{num\_syllables}
\begin{itemize}
  \item Number of syllables in the target word, via the Syllables library.
\end{itemize}
\texttt{is\_acronym}
\begin{itemize}
  \item Boolean for whether the target word is all capital letters.
\end{itemize}

\subsection{Semantic Features}

\texttt{num\_hyperyms}
\begin{itemize}
  \item Number of hyperyms associated with the target word. The target word is initially disambiguated using NLTK's implementation of the Lesk algorithm for Word Sense Disambiguation (WSD) \citep{lesk1986automatic}, which finds the WordNet Synset with the highest number of overlapping words between the context and different definitions of each Synset.
\end{itemize}
\texttt{num\_hyponyms}
\begin{itemize}
  \item Number of hyponyms associated with the target word. Procedure for finding this is analogous to that for \texttt{num\_hyperyms}.
\end{itemize}
\texttt{glove\_word}
\begin{itemize}
  \item 300-dimension embedding for each target word, pre-trained on Wikipedia-2014 and Gigaword. Target word is lowercased for ease.
\end{itemize}
\texttt{elmo\_word}
\begin{itemize}
  \item 1024-dimension embedding for each target word, pre-trained on the One Billion Word Benchmark corpus.
\end{itemize}
\texttt{glove\_context}
\begin{itemize}
  \item 300-dimension average of GloVe word embeddings (see \texttt{glove\_word} above) for each word in the given context. Each word is lowercased for simplicity.
\end{itemize}
\texttt{infersent\_embeddings}
\begin{itemize}
  \item 4096-dimension embedding for the context.
\end{itemize}

\subsection{Phonetic Features}

\texttt{char\_transition\_min}
\begin{itemize}
  \item Minimum of the set of character transition probabilities for each character bigram in the target word. Ground truth character transition probabilities between any two English characters are estimated over Gigaword.
\end{itemize}
\texttt{char\_transition\_max}
\begin{itemize}
  \item Maximum of the set described above.
\end{itemize}
\texttt{char\_transition\_mean}
\begin{itemize}
  \item Mean of the set described above.
\end{itemize}
\texttt{char\_transition\_std}
\begin{itemize}
  \item Standard deviation of the set described above.
\end{itemize}
\texttt{phoneme\_transition\_min}
\begin{itemize}
  \item Minimum of the set of phoneme transition probabilities for each character bigram in the target word. Ground truth phoneme transition probabilities between any two phonemes are estimated over the Gigaword corpus. The phoneme set considered is that of the CMU Pronouncing Dictionary.\footnote{\url{http://speech.cs.cmu.edu/cgi-bin/cmudict}}
\end{itemize}
\texttt{phoneme\_transition\_max}
\begin{itemize}
  \item Maximum of the set described above.
\end{itemize}
\texttt{phoneme\_transition\_mean}
\begin{itemize}
  \item Mean of the set described above.
\end{itemize}
\texttt{phoneme\_transition\_std}
\begin{itemize}
  \item Standard deviation of the set described above.
\end{itemize}

\subsection{Word Frequency \& N-gram Features}

\subsubsection{Gigaword-based}

\texttt{tf}
\begin{itemize}
  \item Target word term frequency. Note that all term frequency-based features are computed using Scikit-learn library's \texttt{CountVectorizer} \citep{pedregosa2011scikit}.
\end{itemize}
\texttt{tf\_lemma}
\begin{itemize}
  \item Term frequency of the lemmatized target word. Lemmatization is performed using NLTK's WordNet Lemmatizer.
\end{itemize}
\texttt{tf\_summed\_bpe}
\begin{itemize}
  \item Sum of term frequencies of each BPE in the target word. BPE tokenization is performed using Hugging Face's BERT Tokenizer.
\end{itemize}
\texttt{tf\_ngram\_2}
\begin{itemize}
  \item Sum of the term frequencies of each bigram in the context containing the target word.
\end{itemize}
\texttt{tf\_ngram\_3}
\begin{itemize}
  \item Sum of the term frequencies of each trigram in the context containing the target word.
\end{itemize}
\texttt{tfidf}
\begin{itemize}
  \item Term frequency-inverse document frequency.
\end{itemize}
\texttt{tfidf\_ngram\_2}
\begin{itemize}
  \item Sum of the term frequency-inverse document frequencies of each bigram in the context containing the target word.
\end{itemize}
\texttt{tfidf\_ngram\_3}
\begin{itemize}
  \item Sum of the term frequency-inverse document frequencies of each trigram in the context containing the target word.
\end{itemize}

\subsubsection{Google N-gram-based}

\texttt{google\_ngram\_1}
\begin{itemize}
  \item Term frequency of the target word.
\end{itemize}
\texttt{google\_ngram\_2\_head}
\begin{itemize}
  \item Term frequency of leading bigram in the context containing the target word.
\end{itemize}
\texttt{google\_ngram\_2\_tail}
\begin{itemize}
  \item Term frequency of trailing bigram in the context containing the target word.
\end{itemize}
\texttt{google\_ngram\_2\_min}
\begin{itemize}
  \item Minimum of the set of term frequencies of bigrams in context containing the target word.
\end{itemize}
\texttt{google\_ngram\_2\_max}
\begin{itemize}
  \item Maximum of the set described above.
\end{itemize}
\texttt{google\_ngram\_2\_mean}
\begin{itemize}
  \item Average of the set described above.
\end{itemize}
\texttt{google\_ngram\_2\_std}
\begin{itemize}
  \item Standard deviation of the set described above.
\end{itemize}
\texttt{google\_ngram\_3\_head}
\begin{itemize}
  \item Term frequency of leading trigram in the context containing the target word.
\end{itemize}
\texttt{google\_ngram\_3\_mid}
\begin{itemize}
  \item Term frequency of middle trigram in the context containing the target word.
\end{itemize}
\texttt{google\_ngram\_3\_tail}
\begin{itemize}
  \item Term frequency of trailing trigram in the context containing the target word.
\end{itemize}
\texttt{google\_ngram\_3\_min}
\begin{itemize}
  \item Minimum of set of term frequencies of trigrams in the context containing target word.
\end{itemize}
\texttt{google\_ngram\_3\_max}
\begin{itemize}
  \item Maximum of the set described above.
\end{itemize}
\texttt{google\_ngram\_3\_mean}
\begin{itemize}
  \item Average of the set described above.
\end{itemize}
\texttt{google\_ngrams\_3\_std}
\begin{itemize}
  \item Standard deviation of the set described above.
\end{itemize}

\subsubsection{SUBTLEXus-based}

\texttt{FREQcount}
\begin{itemize}
  \item Number of times target word appears in corpus.
\end{itemize}
\texttt{CDcount}
\begin{itemize}
  \item Number of films in which target word appears.
\end{itemize}
\texttt{FREQlow}
\begin{itemize}
  \item Number of times the lowercased target word appears in corpus.
\end{itemize}
\texttt{CDlow}
\begin{itemize}
  \item Number of films in which the lowercased target word appears.
\end{itemize}
\texttt{SUBTLWF}
\begin{itemize}
  \item Number of times the target word appears per million words.
\end{itemize}
\texttt{SUBTLCD}
\begin{itemize}
  \item Percent of films in which target word appears.
\end{itemize}

\subsubsection{BNC-based}

\texttt{bnc\_frequency}: Target word term frequency.

\subsection{Syntactic Features}

\texttt{parse\_tree\_depth}
\begin{itemize}
  \item Height of context's constituency parse tree. Parse trees are obtained using a Stanford CoreNLP pipeline.
\end{itemize}
\texttt{token\_depth}
\begin{itemize}
  \item Depth of the target word with respect to root node of the context's constituency parse tree.
\end{itemize}
\texttt{num\_words\_at\_depth}
\begin{itemize}
  \item Number of words at the depth of the target word (see \texttt{token\_depth} above) in the context's constituency parse tree.
\end{itemize}
\texttt{is\_proper}
\begin{itemize}
  \item Boolean for whether target word is a proper noun/adjective, based on capitalization.
\end{itemize}
\texttt{POS\_\{CC, CD, DT, EX, FW, IN, JJ, JJR, JJS, LS, MD, NN, NNP, NNPS, NNS, PDT, POS, PRP, PRP\$, RB, RBR, RBS, RP, SYM, TO, UH, VB, VBD, VBG, VBN, VBP, VBZ, WDT, WP, WP\$, WRB\}}
\begin{itemize}
  \item Booleans indicating the target word's part-of-speech tag. Tags considered are those used in the Penn Treebank Project.\footnote{\url{https://www.ling.upenn.edu/courses/Fall_2003/ling001/penn_treebank_pos.html}} Tags are estimated using NLTK's \texttt{pos\_tag} method.
\end{itemize}

\subsection{Readability Features}

\texttt{automated\_readability\_index, avg\_character\_per\_word, avg\_letter\_per\_word, avg\_syllables\_per\_word, char\_count, coleman\_liau\_index, crawford, fernandez\_huerta, flesch\_kincaid\_grade, flesch\_reading\_ease, gutierrez\_polini, letter\_count, lexicon\_count, linsear\_write\_formula, lix, polysyllabcount, reading\_time, rix, syllable\_count, szigriszt\_pazos, SMOGIndex, DaleChallIndex}
\begin{itemize}
  \item Algorithms applied using Textstat library implementations, most being readability metrics.
\end{itemize}

\subsection{Other Features}

\texttt{ppl}
\begin{itemize}
  \item Perplexity metric, as defined by the Hugging Face library.\footnote{\url{https://huggingface.co/transformers/perplexity.html}} For each token in the context, we use a pre-trained GPT-2 model to estimate the $\log$-likelihood of the token occurring \textit{given its preceding tokens}. A sliding-window approach is used to handle the large number of tokens in a context. The $\log$-likelihoods are averaged, and then exponentiated.
\end{itemize}
\texttt{ppl\_aspect\_only}
\begin{itemize}
  \item Similar approach to that described above, where only $\log$-likelihoods of tokens comprising the target word are averaged.
\end{itemize}
\texttt{num\_OOV}
\begin{itemize}
  \item Number of words in the context that do not exist in the vocabulary of Gigaword.
\end{itemize}
\texttt{corpus\_bible, corpus\_biomed, corpus\_europarl}
\begin{itemize}
  \item Booleans indicating the sample's domain.
\end{itemize}

\section{Model Hyperparameters}
\label{appendix:hyperparameters}

Here we provide optimized hyperparameter settings that may help future developers with reproducing results, namely with training our models.

\subsection{XGBoost}

Below are tuned parameters used for all of our XGBoost models. Parameters not listed are given default values as specified in documentation:\footnote{\url{https://xgboost.readthedocs.io/en/latest/}}

\indent \texttt{colsample\_bytree}: 0.7\\
\indent \texttt{learning\_rate}: 0.03\\
\indent \texttt{max\_depth}: 5\\
\indent \texttt{min\_child\_weight}: 4\\
\indent \texttt{n\_estimators}: 225\\
\indent \texttt{nthread}: 4\\
\indent \texttt{objective}: `reg:linear'\\
\indent \texttt{silent}: 1\\
\indent \texttt{subsample}: 0.7

\subsection{MT-DNN}

MT-DNN uses \texttt{yaml} as its config file format. Below are the contents of our task config file:

\indent \texttt{data\_format}: PremiseAndOneHypothesis\\
\indent \texttt{enable\_san}: false\\
\indent \texttt{metric\_meta}:\\
\indent - Pearson\\
\indent - Spearman\\
\indent \texttt{n\_class}: 1\\
\indent \texttt{loss}: MseCriterion\\
\indent \texttt{kd\_loss}: MseCriterion\\
\indent \texttt{adv\_loss}: MseCriterion\\
\indent \texttt{task\_type}: Regression

\subsection{Ensemble}

Threshold above which a sample is assigned its \textit{reduced} prediction (ie. XGBoost$_\textit{reduced}$ prediction) instead of its \textit{full} prediction (ie. XGBoost$_\textit{full}$ prediction): 0.59. Note that this threshold is used to compute our XGBoost$
_{\textit{full}+\textit{reduced}}$ prediction.\\

\noindent Weighted average ensemble (single word subtask):\\
- Weight for XGBoost$_{\textit{full}+\textit{reduced}}$ prediction: 0.5\\
- Weight for MT-DNN prediction: 0.5\\

\noindent Weighted average ensemble (MWE subtask):\\
- Weight for XGBoost$_{\textit{full}+\textit{reduced}}(\text{head})$: 0.28 \\
- Weight for XGBoost$_{\textit{full}+\textit{reduced}}(\text{tail})$: 0.17 \\
- Weight for MT-DNN prediction: 0.55\\

\end{document}